\documentclass{IEEEtran}

\usepackage{times}
\usepackage{graphicx}
\usepackage{xcolor}

\usepackage[numbers,sort]{natbib}
\usepackage{multicol}
\usepackage[bookmarks=true]{hyperref}

\begin{document}

% paper title
\title{CoHRT: A Collaboration System for Human-Robot Teamwork}

% You will get a Paper-ID when submitting a pdf file to the conference system
% \author{Author Names Omitted for Anonymous Review. Paper-ID [add your ID here]}

\author{\IEEEauthorblockN{Sujan Sarker, Haley N. Green, Mohammad Samin Yasar, Tariq Iqbal}
\\
\IEEEauthorblockA{School of Engineering and Applied Science, University of Virginia, Charlottesville, USA} 
\\
%\textit{University of Virginia}\\
%Charlottesville, USA
%hng9vf@virginia.edu, mi8uu@virginia.edu, sma6zw@virginia.edu, tiqbal@virginia.edu
\{zzr2hs, hng9vf, msy9an, tiqbal\}@virginia.edu
}

\maketitle

\begin{abstract}
Collaborative robots are increasingly deployed alongside humans in factories, hospitals, schools, and other domains to enhance teamwork and efficiency. Systems that seamlessly integrate humans and robots into cohesive teams for coordinated and efficient task execution are needed, enabling studies on how robot collaboration policies affect team performance and teammates' perceived fairness, trust, and safety. Such a system can also be utilized to study the impact of a robot's normative behavior on team collaboration. Additionally, it allows for investigation into how the legibility and predictability of robot actions affect human-robot teamwork and perceived safety and trust. Existing systems are limited, typically involving one human and one robot, and thus require more insight into broader team dynamics. Many rely on games or virtual simulations, neglecting the impact of a robot's physical presence. Most tasks are turn-based, hindering simultaneous execution and affecting efficiency. This paper introduces CoHRT (Collaboration System for Human-Robot Teamwork), which facilitates multi-human-robot teamwork through seamless collaboration, coordination, and communication. CoHRT utilizes a server-client-based architecture, a vision-based system to track task environments, and a simple interface for team action coordination. It allows for the design of tasks considering the human teammates' physical and mental workload and varied skill labels across the team members. We used CoHRT to design a collaborative block manipulation and jigsaw puzzle-solving task in a team of one Franka Emika Panda robot and two humans. The system enables recording multi-modal collaboration data to develop adaptive collaboration policies for robots. To further utilize CoHRT, we outline potential research directions in diverse human-robot collaborative tasks.

\end{abstract}

\IEEEpeerreviewmaketitle

\section{Introduction}
Collaborative robots (cobots) are increasingly being deployed to work alongside humans in various domains, including manufacturing, healthcare, and education \cite{sanneman2021state, Smart, hoffman2019evaluating, sarker2024fold}. These cobots have the potential to revolutionize efficiency and productivity by seamlessly integrating into human-robot teams and facilitating coordinated task execution \cite{iqbal2017coordination, iqbal2019fast, lee2023reimagining,rahman2022ai,sarker2021robotics, admoni2014deliberate}. By sharing physical workspaces and collaborating on complex tasks, cobots can augment human capabilities, alleviate physical and cognitive burdens, and complete intricate objectives that would be challenging for either humans or robots alone to accomplish effectively \cite{Garry, Carmen, gombolay2015coordination, hinds2004whose, kwon2018emotional, iqbal2021temporat_anticipation}. However, realizing the full potential of human-robot collaboration hinges on developing systems that enable efficient teamwork, coordination, and communication between humans and robots \cite{gombolay2017computational, chen2018planning, nikolaidis2017human}. Such a system allows the robot to perceive the human teammates' behavior, understand their instruction, and leverage its collaborative efforts accordingly \cite{islam2023eqa, islam2023patron, islam2022caesar,  islam2023maven, samyoun2022m3sense,  yasar2023vader, yasar2023imprint}.

The development of collaboration systems must not only focus on efficiency and productivity but also prioritize the physical, cognitive, and social safety of human team members \cite{ZACHARAKI2020104667, AKALIN2022102744, RUBAGOTTI2022104047, vanWaveren2023increasing}. This includes designing robots with predictable and legible behaviors, implementing robust safety mechanisms, and considering the influence of social and cultural norms on perceived safety in human-robot interactions \cite{habibian2022encouraging, yang2024enhancing, seraj2024interactive}. Furthermore, as these collaborative systems evolve, it is crucial to explore how they can be designed to enhance trust, fairness, and overall acceptance of robots in shared workspaces \cite{haley2022hri, Jung2020thri, jung2020robots, chang2020defining, chang2021unfair, joosse2017groups}.

\begin{figure}[t] 
     \centering
\includegraphics[width=0.5\textwidth] {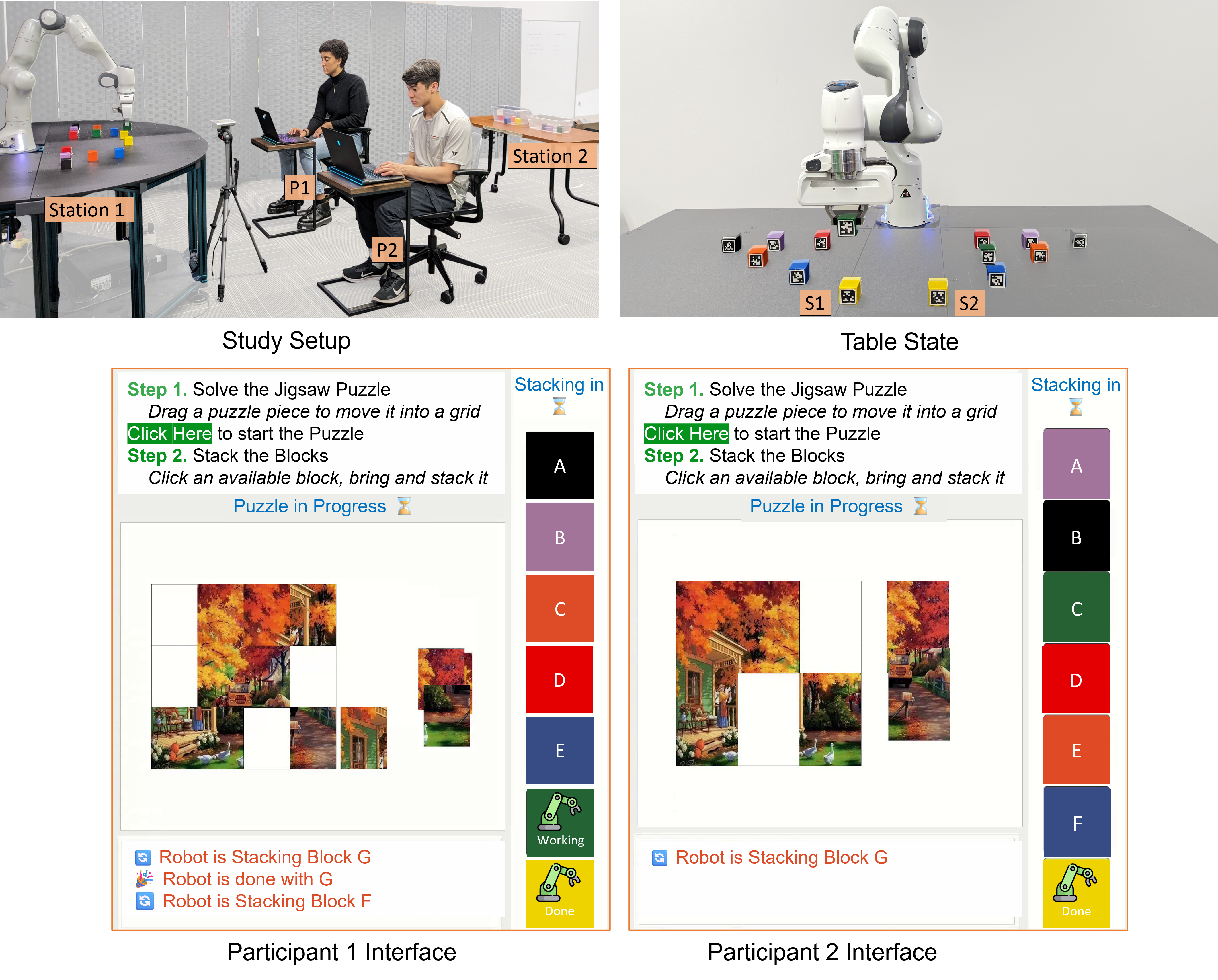}
     \caption{A Franka Emika Panda robot collaborates with a team of two human participants. The task sequence for each participant involves first solving a jigsaw puzzle, which is either $3\times3$ or $3\times2$ in size, followed by stacking seven blocks according to a specified color pattern. The robot's role is limited to assisting the participants exclusively with the block stacking task. A graphical user interface (GUI) displays the current state of the puzzles and the block stacks.}
     \label{fig:study_setup}
\end{figure}

\par Recent research has explored various aspects of human-robot teamwork, such as communication, synchronization, and task allocation \cite{iqbal2015method, iqbal2015joint, iqbal2016movement, shah2011improved, hoffman2010synchronization}. These studies highlight the importance of designing effective collaboration strategies considering human-robot teams' unique dynamics and challenges \cite{sebo2020robots, Guy2019Fluency, nikolaidis2017human}. However, existing human-robot collaboration systems have limitations that hinder the robot's ability to facilitate seamless and efficient teamwork. Many of these systems are confined to dyadic interactions involving only one human and one robot \cite{chang2020defining, chang2020tasc}, failing to capture the dynamics and complexities of larger team settings, where multiple humans and robots must coordinate their actions and adapt to diverse individual strengths, preferences, and constraints. 

\par Additionally, several systems rely on game-based or virtual simulation environments, neglecting the potential impact of a robot's physical embodiment on collaboration dynamics, such as spatial awareness, non-verbal communication, and shared situational understanding \cite{claure2020multi, claure2020ai, claure2024multiplayer, CLAURE2023Machine, mailapalli2022modeling}. Moreover, some works utilize complex systems, such as motion capture systems, to perceive and comprehend human behavior and determine the collaboration effort required by the robot \cite{yasar2024hri}. In such a system, participants must be equipped with sophisticated sensors or apparatus, which may increase participants' cognitive load and discomfort, thus affecting their collaboration experiences. Furthermore, many tasks in these systems are turn-based, restricting simultaneous execution and potentially hindering team efficiency by preventing concurrent action and fluid task handoffs \cite{chang2020defining, Jung2020thri}.

\par To address these limitations, we introduce CoHRT (Collaboration System for Human-Robot Teamwork), a system designed to facilitate multi-human-robot teamwork through seamless collaboration, coordination, and communication. CoHRT leverages a vision-based system for tracking the environment state and team members' actions, enabling real-time monitoring. It also incorporates a simple interface for action coordination among team members, allowing for efficient communication and synchronization of efforts. Importantly, CoHRT allows for the design of tasks that accommodate varying skill levels and constraints across team members, enabling studies on team performance, trust, fairness, and robot acceptance in diverse settings that more accurately reflect real-world scenarios. We demonstrate the capability of the CoHRT system through a collaborative task involving one Franka Emika Panda robot and two human participants. The task requires the team to solve a jigsaw puzzle and stack blocks, creating a scenario that demands a mental and physical workload. 

\par The CoHRT system is designed to be extensible to larger teams and diverse task domains, allowing for the exploration of various team compositions, task complexities, and environmental constraints. By enabling the collection of multi-modal collaboration datasets, CoHRT can facilitate the development of adaptive collaborative policies that optimize team performance, enhance trust, and promote acceptance of robotic teammates. We expect CoHRT to be a valuable resource for the broader human-robot collaboration research community. 

\section{The CoHRT System}
\begin{figure}[!t] 
     \centering
         \includegraphics[width=0.45\textwidth] {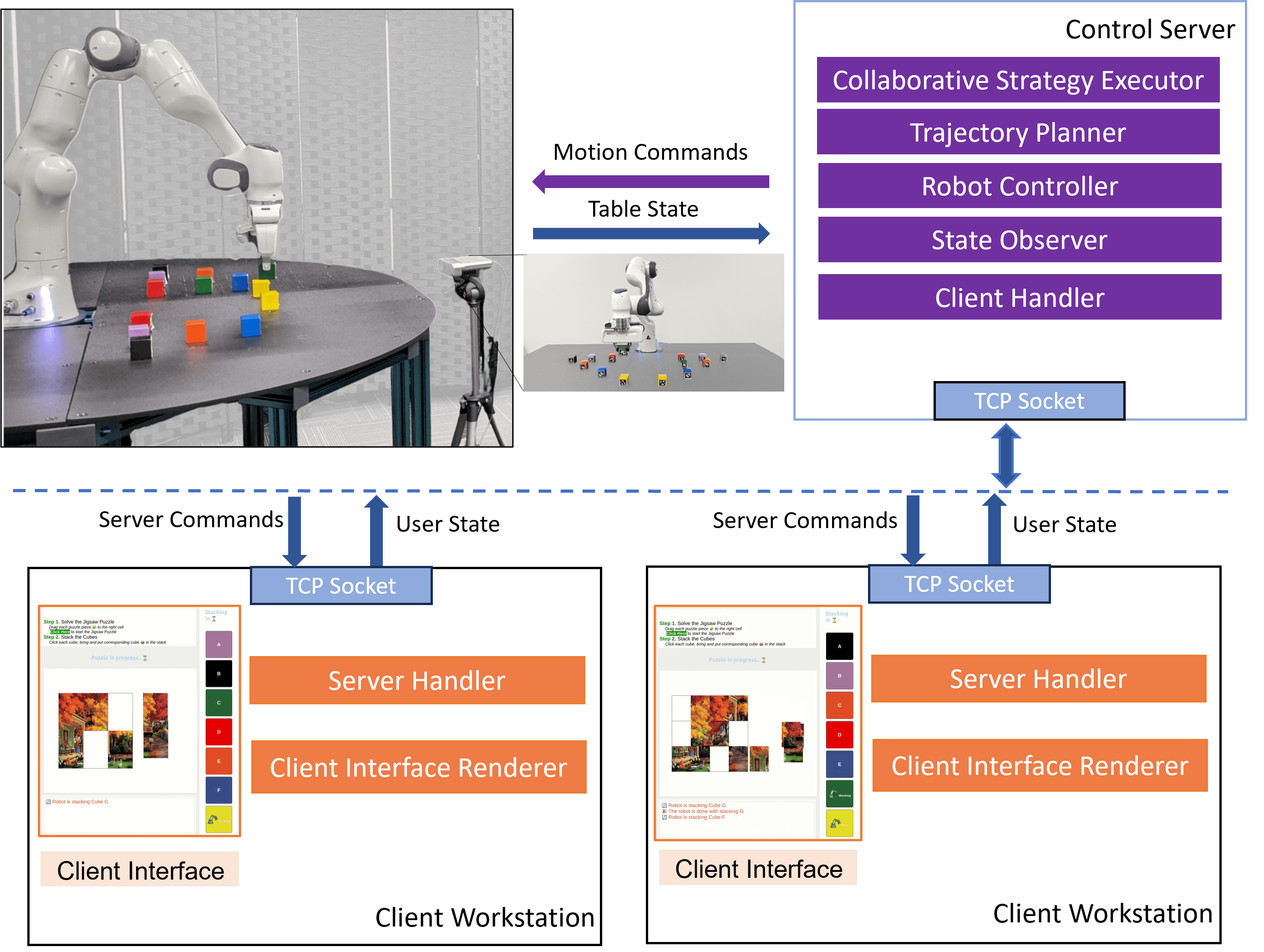}
     \caption{ Architecture of the human-robot collaboration system. The system consists of a server program and client programs communicating via TCP sockets. The server program includes modules for executing collaboration strategies, trajectory planning, robot control, state observation, and client handling. The client program provides a user interface for the puzzle-solving task, renders the interface, and exchanges messages with the server. The system facilitates synchronized coordination between the robot and human teammates during the collaborative task.}
     \label{fig:system_architecture}
\end{figure}

\subsection{The Robot}
The CoHRT system utilizes the Franka Emika Panda robot, a state-of-the-art manipulator robot designed for safe
and efficient interaction with humans in various industrial and research settings. With its seven joints and lightweight design, the Panda robot offers remarkable versatility and agility, enabling it to execute intricate manipulations and adapt to diverse tasks. The robot's sophisticated control software and torque-controlled joints ensure fluid, accurate, and responsive movements. Moreover, the Panda robot prioritizes safety through integrated features like collision detection and force sensing, allowing direct human collaboration without protective barriers. These features, along with its user-friendly interface and programming capabilities, make the Franka Emika Panda robot an ideal choice for studying human-robot collaborative teamwork. The robotic manipulator employs pick-and-place operations, the fundamentals for most mobile manipulators \cite{Jung2020thri}. These basic pick-and-place capabilities can be leveraged to develop various collaborative task scenarios. By maintaining simplicity, the CoHRT system facilitates a clearer understanding of the effect of robot collaborative policy on human teammates.
%%%%
\subsection{The Collaborative Task}
The CoHRT system enables the design of two types of tasks: tasks involving physical workloads, like block manipulation, and tasks requiring mental workload, such as solving puzzles and math problems. By offering this flexibility, the CoHRT system can simulate real-world scenarios, such as assembly tasks frequently employed in human-robot collaboration research \cite{Jung2020thri, hayes2016autonomously}. In this context, the mentally demanding task is comparable to a planning task, while the physically demanding task parallels the plan execution.

\subsection{CoHRT Architecture} 
The CoHRT system leverages a server-client architecture, where a control server acts as the central coordinator, facilitating collaboration and coordination among the human-robot team (Fig. \ref{fig:system_architecture}). This section unfolds different modules of CoHRT architecture. While describing different modules, we consider a scenario where the CoHRT system is utilized to develop a block stacking and a picture jigsaw puzzle-solving task for human teammates. In contrast, the robot's task is limited to block stacking tasks.
\subsubsection{Server Program} The server program is a central controller and facilitates communication and coordination between the robot and participants. The server program also executes robot collaboration strategies and controls the robot by planning the trajectory. We provide a detailed description of each of the modules below.\\
\textbf{Collaboration Strategy Executor} runs a particular collaboration procedure relying on the system state detected by the state observer module. It also communicates with the trajectory planner to plan the trajectory of the current robot task and send it to the robot controller module.\\
\textbf{Trajectory Planner} determines the set of robot waypoints based on the current state by determining the block fetch location and stack height.\\
\textbf{Robot Controller} implements low-level planning, including inverse kinematics (ik) solvers and joint angle calculators. We rely on the Python \textit{frankx} \cite{frankx} library for low-level trajectory planning to move the robot to a target waypoint.\\
\textbf{State Observer} provides the current task state. To detect the puzzle state, we rely on the client program that sends the current puzzle state when the participant takes an action (e.g., moving a puzzle piece to a grid). The stack state changes when the participant or the robot stacks a block. In our system, a block can be in one of three manipulation states: unstacked, working, and stacked. In contrast, each stack can be in one of two states: complete and incomplete. We implement a vision-based approach that detects the current stack state by reading the Apriltags \cite{wang2016iros} attached to each block. 

We utilize an Azure Kinect (RGB-D) \cite{azurekinectdk} camera to capture high-resolution RGB images of the stacks, which are then processed using the Python \textit{pupil-apriltags} library \cite{pupil-apriltags-doc} to detect and decode the tags. The unique identifier of each Apriltag allows us to distinguish individual blocks and their positions within the stack. Their detected positions enable us to determine relative block locations and accurately detect stack configurations, even when blocks are partially occluded. Based on these detections, the system updates the individual block's state (unstacked or stacked) and the stack state (complete or incomplete). When a robot or participant selects a block for manipulation, the system updates its manipulator property (Human or Robot) and sets it to a ``working" state. The system maintains a history of recent states and enforces that only the topmost unstacked block can be chosen for manipulation. Upon stacking, the system detects the block's Apriltag and leverages its position to determine the correct placement. Each Apriltag is unique so that the system can extract block properties such as color. If the system confirms correct stacking, it updates the block's state to ``stacked.'' Finally, the block's states are updated in the client interface, providing a comprehensive and real-time tracking of the task's progress.\\
\textbf{Client Hander} facilitates communication with the participants by sending task configuration at the beginning of the task and exchanging commands between server and client during the tasks.
\subsubsection{Client Program} provides a client interface for facilitating communication between client and server. The client program also implements the picture jigsaw puzzle-solving task. It has the following modules.\\
\textbf{Client Interface} is the graphical user interface (GUI) implemented using the \textit{PyQt5} \cite{pyqt5} library that renders the puzzle-solving task and the current stack state. Participants can interact with this interface using a standard keyboard. The jigsaw puzzle requires participants to move picture pieces between puzzle grids. To stack a block, the participant first selects the block by clicking. The GUI visualizes the current stack state by showing a human or robot icon on each block and text with "working" or "done." Note that when a block is stacked either by the robot or participant, it is automatically detected by the state observer module of the server, and the state is updated in the client interface. This interface provides a way to communicate actions between the robot and participants. The participant can also observe robot actions and infer their actions, but it can be stressful while solving the puzzle; thus, our system also provides a visualization in the client interface.\\
\textbf{Client Interface Renderer} renders the client interface whenever the state of the system changes.\\
\textbf{Server Handler} is responsible for exchanging messages between the server and client program by implementing a read-and-write procedure to read from and write to the server.

\subsubsection{Interaction and Synchronization}
CoHRT implements a synchronized coordination mechanism to prevent conflicts between team members. During the initial setup phase, the client program receives task configurations, including puzzle size, picture, stack size, and color pattern. The task is initiated by the participant using the client interface. The robot collaborates based on the selected strategy when participants initiate the task. The CoHRT server keeps listening to client requests. An allocation request is sent to the server whenever the robot or participant intends to fetch and place a block. If the block is available, the server reserves it, updates the state accordingly, and communicates the update to the client GUI for visualization.

To handle potential conflicts where the robot and a participant request the same block simultaneously, CoHRT implements a locking mechanism that ensures only one request is entertained. Conflicts can arise when the robot and a participant select the same block for manipulation. It is important to note that each participant has a different stack to manipulate, so there is no conflict between the two participants. To deal with the conflict, we identify the critical sections in the program, such as the current stack state, and implement locks to access these critical sections unambiguously and synchronously. Any selection request first comes to the server, and the server's locking mechanism only confirms one request while the other is rejected. 
This mechanism ensures synchronization and coordination among team members, while the client interface ensures that the coordination is legible to the participants. By implementing this approach, CoHRT effectively manages resource allocation and prevents conflicts in block selection between the robot and human participants.
\section{CoHRT Implementation}
\subsection{An example collaborative task}
To evaluate the CoHRT system, we design a collaborative task involving one robot and two humans (P1 and P2) in a team setting (Fig. \ref{fig:study_setup}). The team's objective is to construct two stacks, each consisting of seven blocks arranged in a specific color pattern, and to solve two jigsaw puzzles of varying complexity ($3\times3$ or $3\times2$ grids). Each block in a stack has a unique color. The human participants are assigned two tasks: first, to solve a picture jigsaw puzzle, and second, to stack blocks in a designated area. The robot's role is to collaborate with the participants solely in the stacking task. Note that the participants do not collaborate but can work simultaneously without interrupting one another, which differs from turn-based approaches \cite{Chang2020RO-MAN, claure2020multi, Jung2020thri}. The puzzle-solving task makes it more mentally demanding for the participants. Each participant is randomly assigned to a $3\times3$ or a $3\times2$. To make the task physically demanding, we introduce the manipulation task, where a participant fetches a block of a specific color from another station (Station 2) and places it in their designated stacking location (S1 and S2) at Station 1. The robot is attached to the stacking station (Station 1) and performs the task exclusively by fetching a block from the same station and placing it in either of the participants' stacks. The manipulator is not mobile, so we place two inventories of blocks, one for each participant, at Station 1. However, participants must fetch blocks from Station 2, making the task more challenging. The designed task is analogous to assembly or building tasks commonly used in many human-robot collaborative studies \cite{Jung2020thri, hayes2016autonomously, hayes2015effective, morioka2010new}. The puzzle-solving task can be considered a planning task. This team scenario is common in many real-life applications, such as factory environments, where each worker has a dedicated task that requires planning and execution. Each participant's actions include moving puzzle blocks between grids during the puzzle-solving phase. During the manipulation phase, they pick and place blocks from one station to another or remain idle. The robot's actions involve picking a block and placing it in either the participants' stacks or remaining idle. The robot equally collaborates with both of the participants by alternatively stacking blocks for each of the participants.
\subsection{Collaborative Task Execution}
\begin{figure}[!t]
     \centering
         \includegraphics[width=0.5\textwidth]{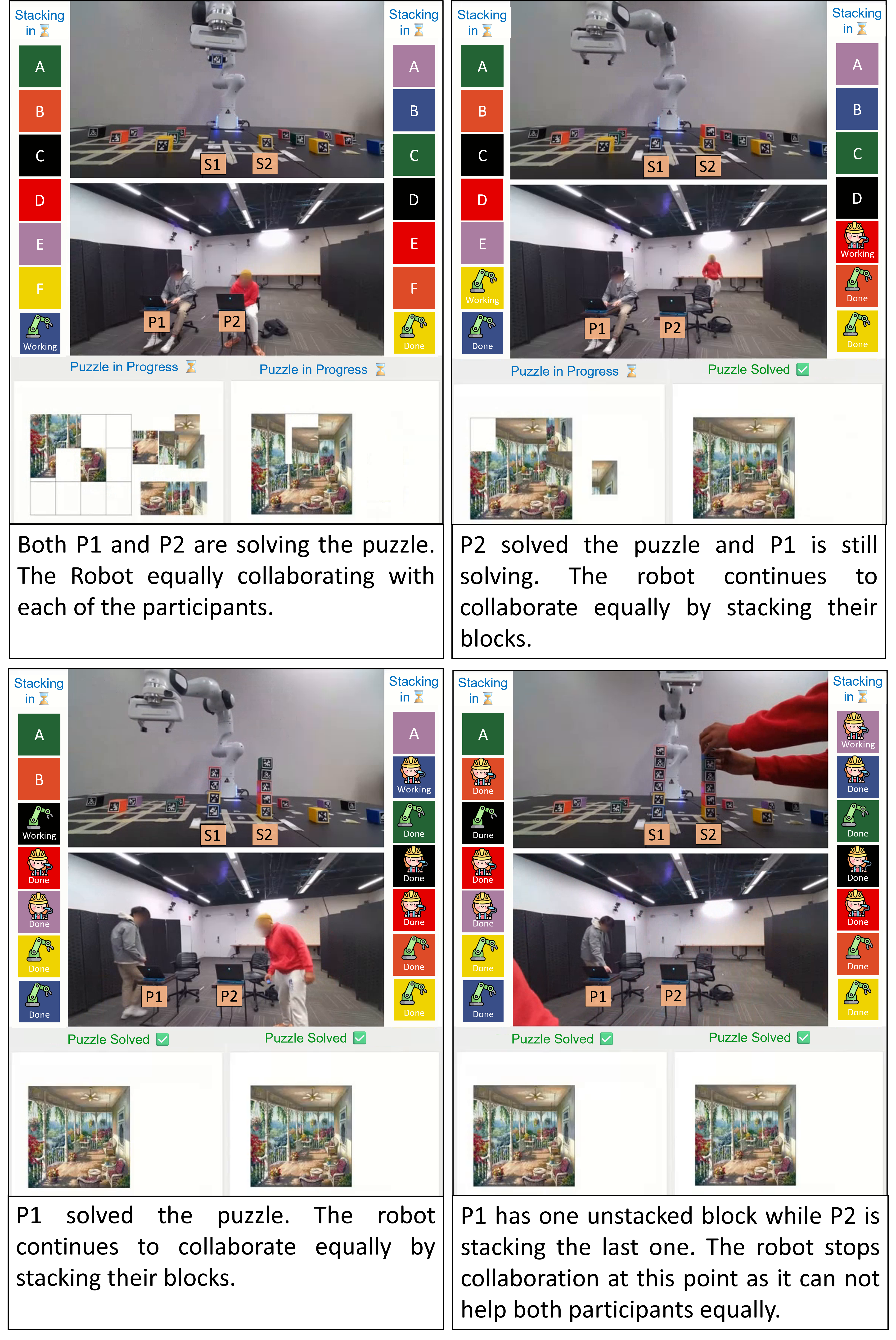}
     \caption{A visual depiction of the interaction scenario between two participants and the Franka Emika Panda robot while performing the collaborative task in the CoHRT system.}
     \label{fig:illustrative_example} 
\end{figure}

Fig. \ref{fig:illustrative_example} shows an illustrative example of the task execution in the human-robot team. The client interface displays the current state of the puzzle and the stack. The participant first solves the jigsaw puzzle by moving the pieces in the interfaces. When the participant finishes solving the puzzle, block-stacking starts. To stack a block, the participant first clicks on the available block within the GUI and then retrieves the physical block from a separate inventory station and places it in the designated location at the stacking station. The robot is attached to the stacking station. It aids the participant by fetching blocks from one of two inventories located at the same station, with each inventory assigned to a specific participant. The robot alternates its collaboration efforts between the participants. The robot continues collaboration till it can provide equal collaboration effort, where equal collaboration means the robot stacks an equal number of blocks for both participants.  The teamwork finished whenever the jigsaw puzzles were solved, and two of the stacks were built.
\section{CoHRT Evaluation Plan}
To assess the effectiveness of the CoHRT system in improving team performance and communication efficiency, we propose a comprehensive evaluation plan incorporating both quantitative and qualitative measures. Our evaluation will focus on team fluency, task performance, and user experience. We will employ several critical metrics from Hoffman's work evaluating fluency in human-robot collaboration \cite{Guy2019Fluency}. The plan includes measuring the team's total time to complete the collaborative task, including puzzle-solving and block-stacking components (task completion time). We will track the duration for which the robot remains inactive while waiting for human actions or decisions (robot idle time) and the periods when human participants are not actively engaged in task-related activities (human idle time). The evaluation will also assess the percentage of time during which both human participants and the robot are simultaneously active (concurrent activity), the time lag between the end of one agent's action and the beginning of another's (functional delay), and the regularity and predictability of action transitions between humans and the robot (human-robot rhythm). By analyzing these metrics, we aim to provide insights into the effectiveness and efficiency of each collaboration strategy in facilitating smooth and seamless human-robot interaction. To evaluate the overall usability of the CoHRT system, we plan to utilize the System Usability Scale (SUS) \cite{brooke1996sus}. Additionally, we will conduct post-task interviews to gather in-depth insights into participants' experiences, challenges, and suggestions for improvement, providing a rich qualitative complement to our quantitative data.
\section{Extendability and Future Work}
The CoHRT shows potential for generalizability and extendability to various task scenarios and team compositions. The server-client-based architecture can be utilized to integrate multiple human participants and robots, enabling the system to scale to larger teams and more complex collaborative tasks. We utilized the CoHRT to design a collaborative task involving two humans and one robot. The system can potentially support collaboration scenarios involving more than two humans by instantiating multiple client interfaces. The CoHRT facilitates seamless collaboration and coordination among the team members and system entities, enabling its extendability to design real-world tasks in manufacturing, healthcare, and education. However, further research and testing would be needed to fully assess the system's capabilities and limitations in the aforementioned scenarios.
In addition to supporting multiple human participants, the system also has the potential to incorporate heterogeneous robots. The CoHRT server can integrate a new robot by instantiating a new robot controller that utilizes robot-specific APIs. This flexibility will enable the system to leverage the capabilities of multiple robots, each with its own set of skills and functionalities, to address more complex and diverse collaborative tasks. For example, a mobile manipulator can be added to the CoHRT to complement the skills of the existing manipulator, providing increased mobility and adaptability to different task environments. A mobile manipulator combines a robotic arm's dexterity and precision with a mobile base's mobility, enabling the robot to navigate and interact with its surroundings more effectively. This adaptability is particularly valuable when the collaborative task requires the robot to move between different workstations or adapt to changing task demands.
\par The system's adaptability also extends to the design of collaborative tasks. While we evaluate CoHRT with a collaborative block stacking task, we can utilize the system to create new tasks such as item sorting and table decluttering tasks. By adjusting the task parameters, perception algorithms, and robot control strategies, the system can be tailored to address specific task requirements and objectives. This allows researchers to explore a variety of collaborative scenarios and investigate the effectiveness of different robot collaboration strategies in diverse task contexts.
In future work, we will explore the extensibility of the CoHRT in larger teams and diverse tasks. We plan to leverage the CoHRT system and the designed tasks to investigate how humans perceive fairness, trust, and safety in collaborative human-robot teams. An intriguing avenue of exploration involves examining the impact of various robot strategies for allocating collaborative efforts on human teammates' perceptions of fairness and trust. Additionally, we intend to collect interaction data to gain insights into teammates' capabilities, which can then be used to develop robot strategies that consider the unique needs of the teammates. By pursuing this line of research, we aim to create more personalized and adaptive robot collaboration strategies that enhance the overall user experience and promote trust and acceptance of human-robot collaboration systems. Furthermore, we will open-source the system for the research community, providing detailed documentation and support to facilitate its adoption in human-robot collaboration research.

%%%
\section{Conclusion} 
\label{sec:conclusion}
In this work, we present CoHRT, a system for seamless human-robot collaboration that addresses the critical limitations of existing systems. CoHRT facilitates multi-human-robot teamwork with synchronized coordination and communication through its server-client architecture and integrated modules. Its flexibility allows for diverse collaborative tasks. By enabling efficient collaboration in mentally and physically demanding tasks, CoHRT opens avenues for research into human-robot team dynamics. The system can be leveraged to study the perception of fairness, trust, and safety in human-robot collaborative tasks, ultimately promoting the development of more user-centric collaboration strategies and the widespread adoption of human-robot collaboration systems.

\bibliographystyle{plainnat}
\bibliography{references}

\end{document}